\documentclass[ review]{elsarticle}

\usepackage[utf8]{inputenc}
\usepackage{graphicx}

\usepackage{lineno,hyperref}
\modulolinenumbers[1]
\usepackage{booktabs}

\usepackage{subfig}

\usepackage{amsmath}

\usepackage[hyperpageref]{backref} 
 \biboptions{sort,compress,super}

\usepackage{multirow}
\usepackage{graphicx}

\usepackage{etoolbox}
\patchcmd{\pprintMaketitle}
  {\fi\hrule}
  {\fi\ifvoid\extrainfobox\else\unvbox\extrainfobox\par\vskip10pt\fi\hrule}
  {}{}

\newenvironment{extrainfo}
  {\global\setbox\extrainfobox=\vbox\bgroup\parindent=0pt }
  {\egroup}
\newsavebox\extrainfobox

\begin{document}

\begin{frontmatter}

\title{Automated Detection of Patellofemoral Osteoarthritis from Knee Lateral View Radiographs Using Deep Learning: Data from the Multicenter Osteoarthritis Study (MOST)}

\author[1]{Neslihan Bayramoglu\corref{cor1}%
\fnref{fn1}}
\ead{firstname.lastname@oulu.fi}
\author[1,2,3]{Miika T. Nieminen}
\author[1,2,3]{Simo Saarakkala}


\begin{extrainfo}
\textbf{Running title:} Deep CNN Patellofemoral Osteoarthitis
\newpage
\end{extrainfo}

\cortext[cor1]{Corresponding author}
\fntext[fn1]{POB 5000, FI-90014, Oulu, Finland}

\address[1]{Research Unit of Medical Imaging, Physics and Technology, University of Oulu, Finland}
\address[2]{Department of Diagnostic Radiology, Oulu University Hospital, Oulu, Finland}
\address[3]{Medical Research Center, University of Oulu and Oulu University Hospital, Oulu, Finland}

\begin{abstract}

\paragraph{Objective}
To assess the ability of imaging-based deep learning to predict radiographic patello\-femoral osteoarthritis (PFOA) from knee lateral view radiographs.

\paragraph{Design}
Knee lateral view radiographs were extracted from The Multicenter Osteoarthritis Study (MOST) (n = 18,436 knees). Patellar region-of-interest (ROI) was first automatically detected, and subsequently, end-to-end deep convolutional neural networks (CNNs) were trained and validated to detect the status of patellofemoral OA. Patellar ROI was detected using deep-learning-based object detection method. Manual PFOA  status  assessment  provided  in the MOST  dataset was used as a classification outcome for the CNNs. Performance of prediction models was assessed using the area under the receiver operating characteristic curve (ROC AUC) and the average precision (AP) obtained from the precision-recall (PR) curve in the stratified 5-fold cross validation setting.
\paragraph{Results}
Of the 18,436 knees, 3,425 (19\%) had PFOA.
AUC and AP for the reference model including age, sex, body mass index (BMI), the total Western Ontario and McMaster Universities Arthritis Index (WOMAC) score, and tibiofemoral Kellgren–Lawrence (KL) grade to predict PFOA were 0.806 and 0.478, respectively.
The CNN model that used only image data significantly improved the prediction of PFOA status (ROC AUC= 0.958, AP= 0.862).

\paragraph{Conclusion}

We present the first machine learning based automatic PFOA detection method. Furthermore, our deep learning based model trained on patella region from knee lateral view radiographs performs better at predicting PFOA than models based on patient characteristics and clinical assessments.

\end{abstract}


 \end{frontmatter}

\section{Introduction}

Plain radiography is commonly used in diagnostics of osteoarthritis (OA) because it is cheap, fast, and widely available.
Both clinical practice and the majority of research studies in OA have traditionally concentrated on the tibiofemoral (TF) joint.
Frontal plane radiography (postero-anterior (PA) view) is routinely used to evaluate the tibiofemoral joint.
However, patellofemoral (PF) joint is the most frequently affected compartment by OA and yet it often remains unrecognized \cite{hinman2007patellofemoral}.
Moreover, patellofemoral osteoarthritis (PFOA) is both highly prevalent \cite{duncan2006prevalence,hart2017prevalence} and  clinically important because it is more strongly associated with knee OA symptoms than tibiofemoral OA \cite{culvenor2014patellofemoral}.
PFOA can occur in the absence of tibiofemoral OA and also in conjunction with it \cite{kobayashi2016prevalence}.
Actually, some studies suggest that OA is more likely to start in the patellofemoral joint and only then extend to the tibiofemoral joint \cite{lankhorst2017incidence, stefanik2016changes, duncan2011incidence}.
Several studies have found that radiographic PFOA cannot be identified using only subject's characteristics and clinical assessments. Therefore, imaging data is needed for diagnosis of PFOA \cite{tan2020can, stefanik2018diagnostic,van2018international, peat2012clinical}.
However, the patellofemoral joint cannot be evaluated from the most commonly used frontal plane radiography.
Consequently, previous studies suggested that PF joint should routinely be considered in knee OA studies by obtaining multiple radiographic views of the knee \cite{zhang2010eular, cicuttini1996choosing};
otherwise  4–7\% of OA cases would be missed \cite{chaisson2000detecting}.

Patellofemoral OA may be an indicator for an early disease process and therefore a possible target for early intervention \cite{duncan2011incidence, macri2017patellofemoral}, which is one of the high priority research areas of The European
League Against Rheumatism (EULAR) \cite{zhang2010eular}.
Additionally, identifying PFOA is important for surgical approaches and rehabilitative treatments \cite{felson2016challenges}.
However, there is a lack of consistency among researchers and clinicians in grading the patellofemoral joint OA status \cite{kobayashi2016prevalence}.
Since clinical features cannot be used to diagnose PFOA \cite{tan2020can,stefanik2018diagnostic,van2018international,peat2012clinical} clear diagnostic guidelines are missing \cite{felson2016challenges}.
Thus,  an accurate analysis of PFOA from imaging data is of high importance to better understand OA and especially its early stages.  
In the current study, we describe the first fully automated method to detect PFOA from lateral view plain radiographs using deep learning.
Our final aim was to compare whether deep convolutional neural network (CNN) features can better identify the presence of definite radiographic PFOA - compared with more conventional machine learning model using only participant characteristics and clinical
assessments.

\begin{figure}[!t]

\includegraphics[width=1.0\linewidth]{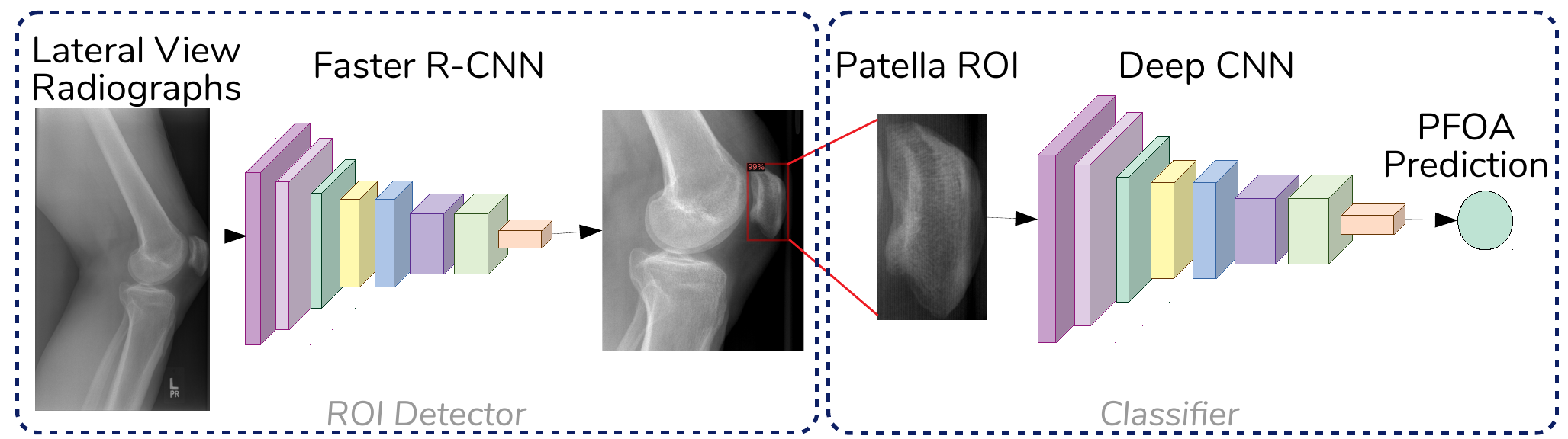}
\caption{Schematic description of the workflow of the current study. We first use a Deep Convolutional Neural Network (CNN)- Faster R-CNN, pretrained on a generic image dataset to detect patellar region-of-interest (ROI) from plain lateral view radiographs.
Subsequently, we used the patellar ROI as the input to the second deep CNN model which was trained from the scratch to predict the patellofemoral osteoarthritis (PFOA) status of the knee. }
\label{fig: pipeline}
\end{figure}

\section{Methods and Materials}

\begin{figure}[!t]
\includegraphics[width=0.7\linewidth]{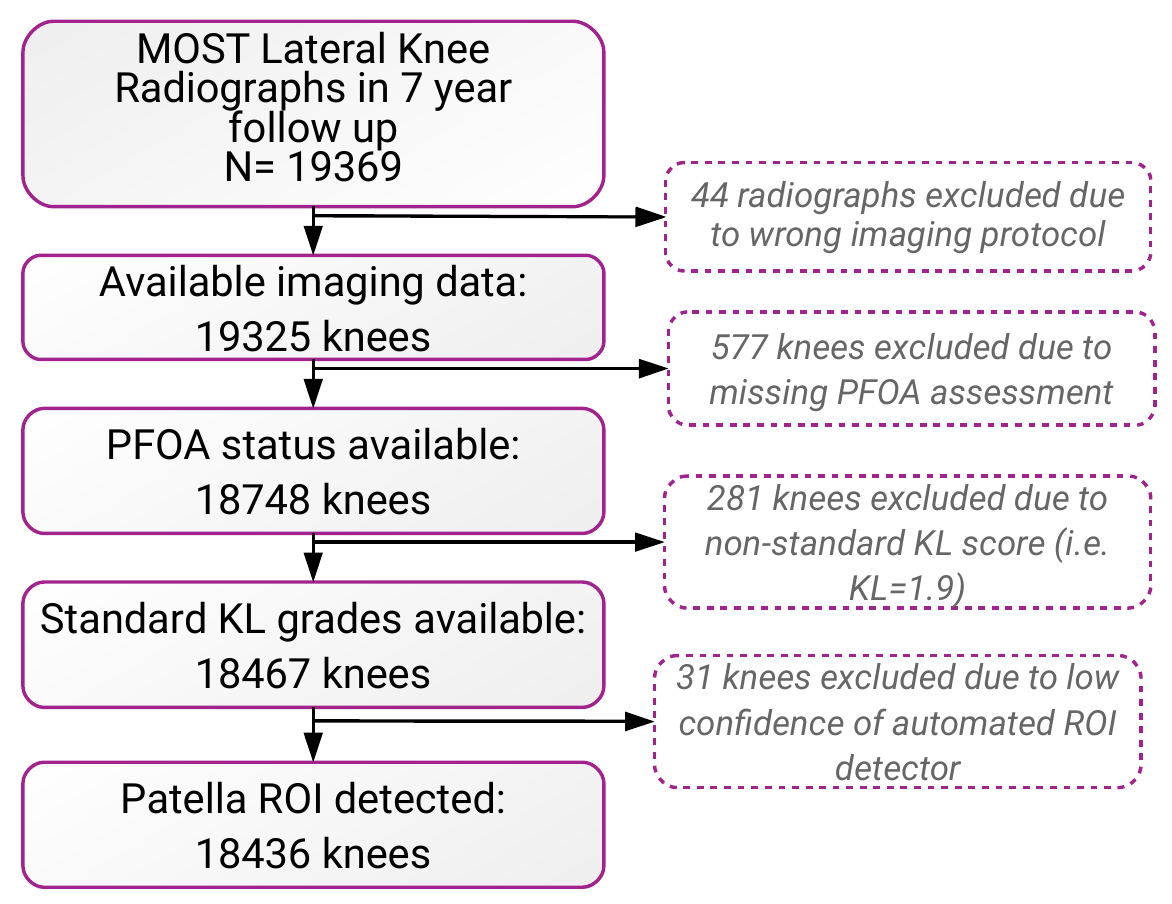}
\caption{Flowchart of the selection of MOST subjects and patellar ROI for the current study. Knees with missing data (radiographs, PFOA status), non-standard Kellgren and Lawrence (KL) scores, and low confidence of patella ROI detector were excluded.}
\label{fig: selection}
\end{figure}

\subsection{Data}
Data from the Multicenter Osteoarthritis Study (MOST, \href{http://most.ucsf.edu}{http://most.ucsf.edu}) was used in this study.
The MOST study is a longitudinal observational study of adults who have or are at high risk
for knee OA.
At baseline, there were 3,026 individuals aged 50–79 years who either had radiographic knee OA or were at high risk for developing the disease.
Knee radiographs were read from the baseline to 15, 30, 60 and 84-month follow-up visits.

Knee joint lateral view radiographs and clinical data at the baseline and follow-up visits were used in the current study. 
In MOST, semiflexed lateral view radiographs were acquired according to a standardized protocol.
Knees with missing data (radiographs, PFOA status), non-standard Kellgren and Lawrence (KL) scores of tibiofemoral joint, and low confidence of patella region of interest (ROI) detector were excluded (Figure \ref{fig: selection}). 
As such, the final subset for assessing PFOA status within the period from baseline to 84 months included 18,436 knees from 2,803 subjects (Table \ref{describe}).

\begin{table}
\small
{\renewcommand{\arraystretch}{0.7}
\caption{Characteristics of the subjects included in the current study. 
BMI: body mass index, WOMAC: the total Western Ontario and McMaster Universities Arthritis Index score, KL: Kellgren and Lawrence scores of tibiofemoral joint, PFOA: patellofemoral osteoarthritis. }
\label{describe}
\begin{center}

\begin{tabular}{llrr}
\toprule
   &  &            non-PFOA &       PFOA \\
\midrule
AGE & mean &     64.7 &    66.2 \\
   & std &      8.3 &     8.1 \\
   & min &     50.0 &    50.0 \\
   & 25\% &     58.0 &    60.0 \\
   & 50\% &     64.0 &    66.0 \\
   & 75\% &     71.0 &    72.0 \\
   & max &     86.0 &    86.0 \\
   \hline
BMI &  mean &     30.0 &    32.7 \\
   & std &      5.5 &     6.6 \\
   & min &     16.0 &    19.7 \\
   & 25\% &     26.2 &    27.9 \\
   & 50\% &     29.3 &    31.7 \\
   & 75\% &     32.9 &    36.2 \\
   & max &     62.4 &    66.1 \\
   \hline
WOMAC &  mean &     16.6 &    27.6 \\
   & std &     16.0 &    17.8 \\
   & min &      0.0 &     0.0 \\
   & 25\% &      3.0 &    14.0 \\
   & 50\% &     12.0 &    26.0 \\
   & 75\% &     26.0 &    39.6 \\
   & max &     91.0 &    92.0 \\
   \hline
KL & mean &      1.1 &     2.5 \\
   & std &      1.3 &     1.2 \\
   & min &      0.0 &     0.0 \\
   & 25\% &      0.0 &     2.0 \\
   & 50\% &      1.0 &     3.0 \\
   & 75\% &      2.0 &     3.0 \\
   & max &      4.0 &     4.0 \\
\bottomrule
\end{tabular}

\end{center}
}
\end{table}

\subsection{Automatic Detection of Patellar Region-of-Interest}

Prior to extraction of the ROIs, the 16-bit DICOM images were normalized using global contrast normalisation and a histogram truncation between the $5^{th}$ and $99^{th}$ percentiles.
These images were eventually converted to 8-bit images ($0-255$ grayscale range).
The image spatial resolution, which was not standardized in the database, was now standardized to $0.2$ mm using a bicubic interpolation.
Right knee images were then horizontally flipped to match the left knee orientation.

State-of-the-art CNN-based object detection algorithm based on a Faster R-CNN design \cite{ren2015faster, wu2019detectron2} was used to automatically detect the patellar ROI from lateral view radiographs. 
596 knee radiographs were manually annotated to train the model. 
We used rectangular ROIs to cover patella. 
We initialized the weights from backbone models pre-trained on COCO (a large-scale object detection) dataset \cite{lin2014microsoft}.
This design was quite effective in predicting rectangular patellar ROI with high confidence (Supplementary Figure \ref{fig: supp_regions} and \ref{fig: supp_regions2}).
By setting a minimum threshold of 90\% certainty, only 31 patellar ROI out of 18,467 knees (0.17 \%) were missed.

\subsection{Predicting Patellofemoral Osteoarthritis Status Using Deep CNN}
We used patellar ROI for predicting the PFOA status using a second deep CNN (Figure \ref{fig: pipeline}).
Our CNN model consists of 3 convolutional layers.
Each  convolution layer (stride= 1, padding= 1) is followed by Batch normalization (BN), max pooling ($2\times2$) and ReLU.
We used two fully connected layers to make the prediction.
A dropout of 0.5 is inserted after the first fully connected layer.

We trained the models from the scratch (end-to-end) using the random weight initialization.
Pre-trained models were not utilized due to custom size of our input (patellar ROI).
An input image size of $128\times64$ was utilized, and we adopted stochastic gradient descent training on a GPU. 
A mini-batch of 64 images were employed, and a momentum of 0.9 was used and trained without weight decay. 
A starting learning rate of 0.001 was first used and decreased by 10 every 8 epochs. 
The models were trained for 20 epochs.

\subsection{Reference Models}
We compared our CNN method with more conventional machine learning based prediction models using the clinical data.
The reference methods were built using Gradient Boosting Machine (GBM) classifier based on decision
tree algorithms \cite{friedman2001greedy} to predict PFOA from clinical data including age, sex, body mass index (BMI), the total Western Ontario and McMaster Universities Arthritis Index (WOMAC) score, and the KL grade of the tibiofemoral joint. 
Following three models were used:
\begin{itemize}
\itemsep0em 
    \item Model 1: Age, Sex, BMI
    \item Model 2: Age, Sex, BMI, total WOMAC score
    \item Model 3: Age, Sex, BMI, total WOMAC score, tibiofemoral OA stage (KL score)
\end{itemize}
GBM models have ability to handle missing values and efficient management of high-dimensional data.
We used LightGBM library \cite{ke2017lightgbm} for the GBM model and the hyperopt package \cite{bergstra2013hyperopt} to find the optimal parameters of the models.
We also analysed the feature importance of the reference models using SHAP library \cite{NIPS2017_7062}.

\subsection{Statistical Analyses}

In order to obtain unbiased estimation of future performance, subject-wise stratified 5-fold cross validation was performed in our experiments. 
Classical k-fold cross-validation often relies on a random partitioning  data into k equal-sized folds.
On the other hand, stratified k-fold cross validation ensures the same distribution of PFOA and non-PFOA cases in both train and validation splits for each fold. 
In addition, we used patient-wise splitting instead of record-wise splitting.
In order to evaluate the performance of the models, out-of-fold (oof) predictions were used. 
We used Receiver Operating Characteristics (ROC) curves and Precision-Recall (PR) curves to assess the performance of the models.
The area under the ROC curve (AUC) and the PR curve (Average Precision, AP) gives summary of the information of these curves. 
PR curves and corresponding AP measures can better highlight performance differences particularly for imbalanced data sets. 
The 95\% confidence intervals were estimated using the stratified bootstrapping with 2000 iterations.
\section{Results}
\noindent\textbf{Cross-validation Results}

Of the 18,436 knees, 3425 (19\%) had PFOA based on the metadata and the PFOA status assessment provided in MOST dataset. 
Radiographic PFOA is defined as follows: Osteophyte score $\geq$ 2 or the joint space narrowing (JSN) score is $\geq1$ plus any osteophyte, sclerosis or cysts $\geq1$ in the PF joint (grades 0–3; 0=normal, 1=mild, 2=moderate, 3=severe).
Individual radiographic features in the MOST dataset were  graded based on the atlases from the Osteoarthritis Research Society International (OARSI) \cite{altman2007atlas} which refers to the previous OARSI atlas for the patellofemoral joint\cite{altman1995atlas} and Framingham Osteoarthritis Study\cite{chaisson2000detecting}.

\begin{figure}
    \centering
    \includegraphics[width=0.7\textwidth]{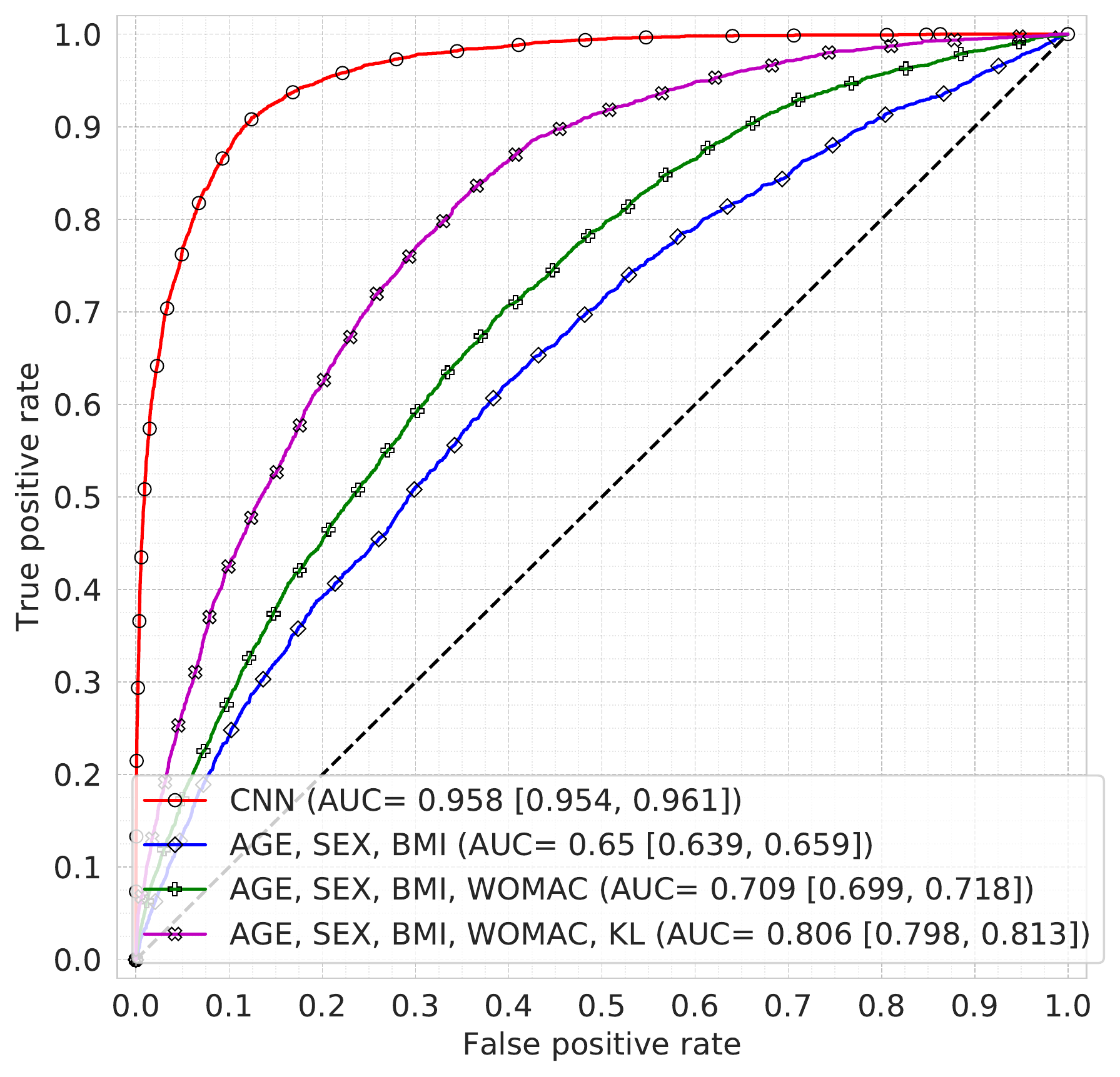}
    \includegraphics[width=0.7\textwidth]{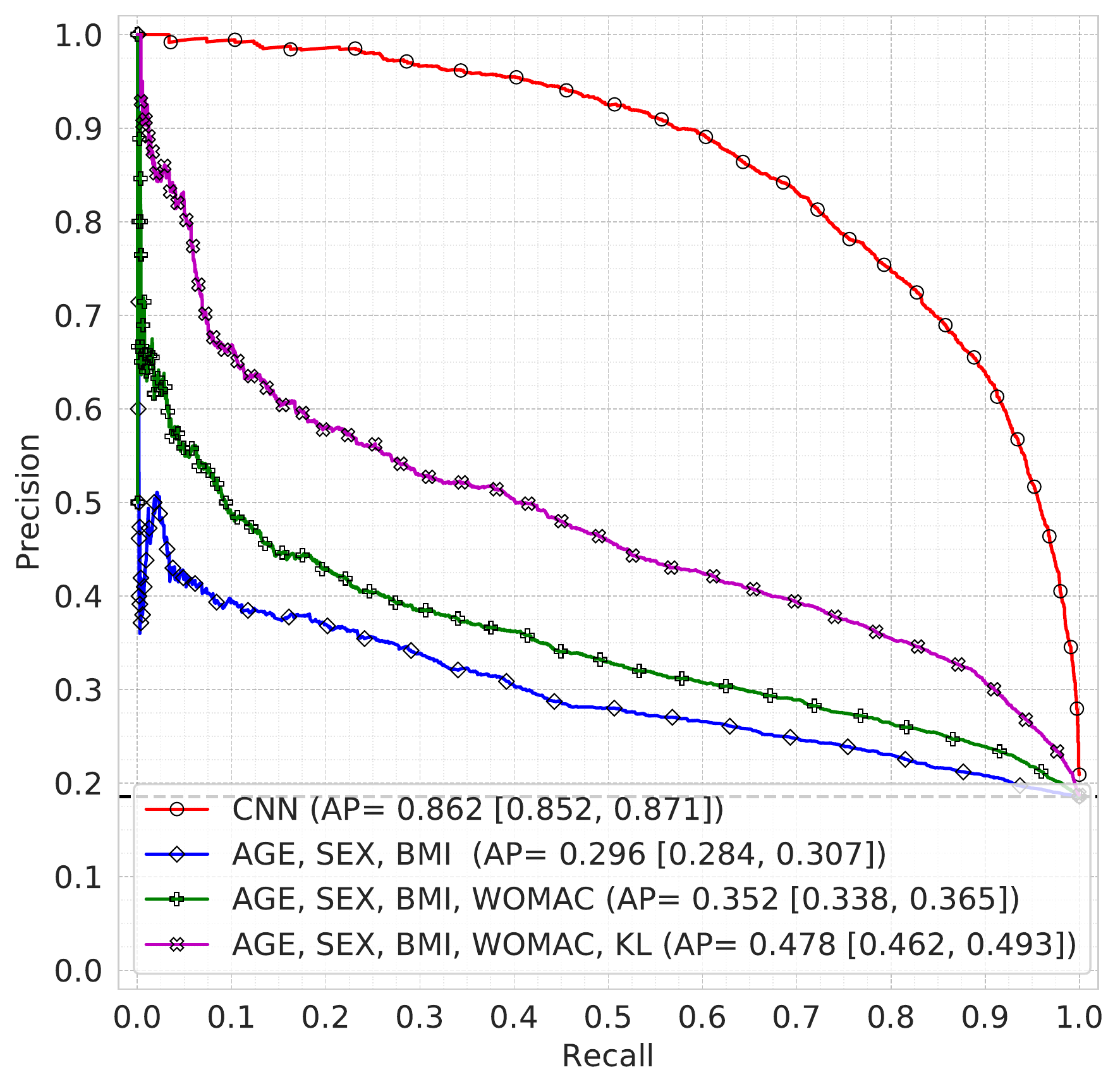}
    \caption{ (a) ROC and (b) PR curves demonstrating the performance of the models. 
    Area under the curves and 95\% confidence intervals in parentheses were given based on a 5-fold cross validation setting.
    Dashed lines in ROC indicate the performance of a random classifier and in case of PR it indicates the distributions of the labels of the dataset (PFOA vs non-PFOA).}
    \label{fig:my_label}
\end{figure}

The performances of all the models are summarized in Figure \ref{fig:my_label}.
Among the performances of the reference models, the model that included age, sex, BMI, total WOMAC score, and KL (model 3) had the highest ROC AUC (0.806 [95\% CI: 0.798–0.813]) and AP score (0.478 [95\% CI: 0.462–0.493]) in the 5-fold cross-validation setting.
ROC AUC values were 0.709 [95\% CI: 0.699–0.718], and 0.65 [95\% CI: 0.639–0.659], AP values were 0.352 [95\% CI: 0.338–0.365] and 0.296 [95\% CI: 0.284–0.307] for model 1 and model 2, respectively.

Our CNN model for predicting the radiographic PFOA in the 5-fold cross validation setting showed the best performance with ROC AUC 0.958 [95\% CI: 0.954–0.961] and AP 0.862 [95\% CI: 0.852–0.871]. 
We compared the CNN model to the strongest reference method model 3 (Figure \ref{fig:my_label}).
We obtained a statistically significant performance difference in AUC (DeLong’s p-value $<$ 1e-10).

\vspace{0.5cm}
\newpage
\noindent\textbf{Evaluation of the CNN Model Performance among Different Groups}

In addition to the cross validation results, evaluations of the CNN model were performed with respect to 
\begin{itemize}
\itemsep0em 
    \item the stages of tibiofemoral OA status (No OA : KL0 and KL1, Early OA: KL2, Severe OA : KL3 and KL4 )
    \item the WOMAC pain scores (Low pain: WOMAC pain score $\leq$ $25^{th}$ percentile, Moderate Pain:  WOMAC pain score between $25^{th} - 75^th$ percentile, High pain: $>$ $75^{th}$ percentile).
\end{itemize}
These results are shown in Table \ref{tab:groups} and in Supplementary Figure \ref{fig: supp_groups}.

\begin{table}[!t]
\caption{Performance of the CNN model with respect to the stages of tibiofemoral OA based on Kellgren-Lawrence (KL) grading (No OA : KL0 and KL1, Early OA: KL2, Severe OA : KL3 and KL4 ) and the total Western Ontario and McMaster Universities Arthritis Index (WOMAC) Pain score (Low pain: WOMAC pain score $\leq$ $25^{th}$ percentile, Moderate Pain:  WOMAC pain score between $25^{th} - 75^th$ percentile, High pain: $>$ $75^{th}$ percentile). AUC and AP indicate the area under the Receiver Operating Characteristics (ROC) curve and Precison-Recall (PR) curves, respectively. }
\label{tab:groups}
\begin{tabular}{llll}
\toprule
                                        &               & AUC {[}95\% CI{]}       & AP {[}95\% CI{]}         \\
                                        \toprule
\multicolumn{1}{c}{\multirow{3}{*}{KL}} & No TF OA      & 0.962 {[}0.954-0.968{]} & 0.715{[}0.678-0.746{]}   \\
\multicolumn{1}{c}{}                    & Early TF OA   & 0.959 {[}0.952-0.965{]} & 0.927 {[}0.916-0.937{]}  \\
\multicolumn{1}{c}{}                    & Severe TF OA  & 0.908 {[}0.899-0.916{]} & 0.864 {[}0.851-0.875{]}  \\
\midrule
\multirow{3}{*}{WOMAC Pain}             & No Pain       & 0.967 {[}0.96-0973{]}   & 0.77 {[}0.735, 0.8{]}    \\
                                        & Moderate Pain & 0.952 {[}0.947-0.957{]} & 0.862 {[}0.849, 0.875{]} \\
                                        & Severe Pain   & 0.943 {[}0.935-0.949{]} & 0.893 {[}0.88, 0.905{]} \\
                                        
\bottomrule

\end{tabular}
\end{table}

\vspace{0.5cm}
\noindent\textbf{Multi-modal Model: Combination of CNN Model, Clinical Features and Patient Characteristics}

We also developed a GBM model that combine the predictions of the CNN model - the probability of PFOA - with age, sex, BMI, WOMAC, and KL grade. 
We used the same 5-fold stratified cross validation setup.
However, this fusion did not introduce any performance increase compared to the image-based CNN model alone (Supplementary Figure \ref{fig: supp_multimodal}).

\section{Discussion}

In this study, we developed a deep learning method for the assessment of radiographic patellofemoral OA status from knee lateral view radiographs and assessed the ability of deep imaging features to predict PFOA.
The trained models were evaluated in patient-wise stratified cross validation setting to assess its robustness.
The discriminative ability of the final model was high (AUC 0.958).

Given the high prevalence of PFOA \cite{duncan2006prevalence,hart2017prevalence}, there is a need to consider also the patellofemoral joint in knee OA research and clinical settings.
In the earlier literature, prediction models based on clinical features and patient characteristics were studied for radiographic PFOA \cite{tan2020can, stefanik2018diagnostic,van2018international, peat2012clinical}.
They all have the same conclusion that confident diagnosis of radiographic PFOA is not possible with clinical signs and patient characteristics alone, and thus, imaging is necessary to confirm diagnosis.
From the experiments, we also found that the diagnostic accuracy of such models were only modest.
To the best of our knowledge, this is the first study to automatically detect radiographic PFOA from imaging data.
Therefore, we believe that it adds a new tool for early OA diagnostics since the disease often starts from the patellofemoral joint \cite{lankhorst2017incidence, stefanik2016changes, duncan2011incidence}.

To assess the potential bias of the trained CNN model, we stratified the data according to the stages of tibiofemoral OA (KL grade) and pain level (WOMAC).
ROC AUC values among different groups were similar, whereas AP score increases with severe pain.
This could be an indication that there is an association of high pain and PFOA, which has been reported previously \cite{cicuttini1996association,lanyon1998radiographic,duncan2009does}, and our CNN model captures some of the symptom-related features from the image data. It is also notable that the combination of patient characteristics and clinical features did not improve the CNN model's performance further, and the feature importance analysis of the multi-modal model (Supplementary Figure \ref{fig: supp_shap}) showed that CNN’s image-based predictions had the strongest impact onto the output.

Major limitation of this study is that we used the MOST (Multi-center Osteoarthritis study) data alone. It is well known that the generalizability of our approach would have been better if we could have utilized two independent datasets for training and testing. Since PFOA is largely unrecognised there is a lack of available data sets that allow the evaluation of PFOA detection from lateral (or skyline) view radiographs. Therefore, we had to use the stratified cross-validation setup with the MOST data. Another limitation of the study is model explanations.
While we are providing the first results in automatic radiographic PFOA prediction from imaging data, we did not provide a “understanding” to characterize the CNN model's ``black box" behavior.
Despite the efforts like attention maps \cite{selvaraju2017grad}, this post-hoc visualization method does not explain the reasoning process of how a network actually makes its decisions.
Therefore, further work is needed for interpreting a deep neural network model and explaining its
predictions in the context of PFOA. 

In conclusion, this study demonstrated the first results for automatic detection of radiographic PFOA from knee lateral view radiographs using deep learning.
Our model had superior discriminative ability over models using patient characteristics and clinical assessments.
Our model could be valuable when building prediction tools for early OA, for surgical approaches, and for rehabilitative treatments.

\section*{\small{Acknowledgments}}
Multicenter Osteoarthritis Study (MOST) Funding Acknowledgment. MOST is comprised of four cooperative grants (Felson – AG18820; Torner – AG18832, Lewis – AG18947, and Nevitt – AG19069) funded by the National Institutes of Health, a branch
of the Department of Health and Human Services, and conducted by MOST study investigators. This manuscript was prepared using MOST data and does not necessarily reflect the opinions or views of MOST investigators.

We gratefully acknowledge the strategic funding of the University of Oulu, Infotech Oulu and the support of NVIDIA Corporation with the donation of the Quadro P6000 GPU used in this research.

\section*{\small{Author Contributions}}
 N.B., M.T.N, and S.S. originated the idea of the study. N.B. performed
the experiments and took major part in writing of the manuscript. M.T.N, and S.S. supervised the project. All  authors participated in producing the final manuscript draft and approved the final submitted version.

\section*{\small{Role of the funding source}}
Funding sources are not associated with the scientific contents of the study.

\section*{\small{Conflict of interest}}
The authors report no conflicts of interest.

\bibliography{bib.bib}

\newpage
\clearpage
\makeatletter
\makeatother


\newcommand{\beginsupplement}{%
        \setcounter{table}{0}
        \renewcommand{\thetable}{S\arabic{table}}%
        \setcounter{figure}{0}
        \renewcommand{\thefigure}{S\arabic{figure}}%
     }

 \beginsupplement
 
 \section*{Supplementary}\label{sec:supp}

\begin{figure}[!ht]
\subfloat[]{\includegraphics[width = 0.7\linewidth, trim={5cm 12cm 8cm 12cm},clip ]{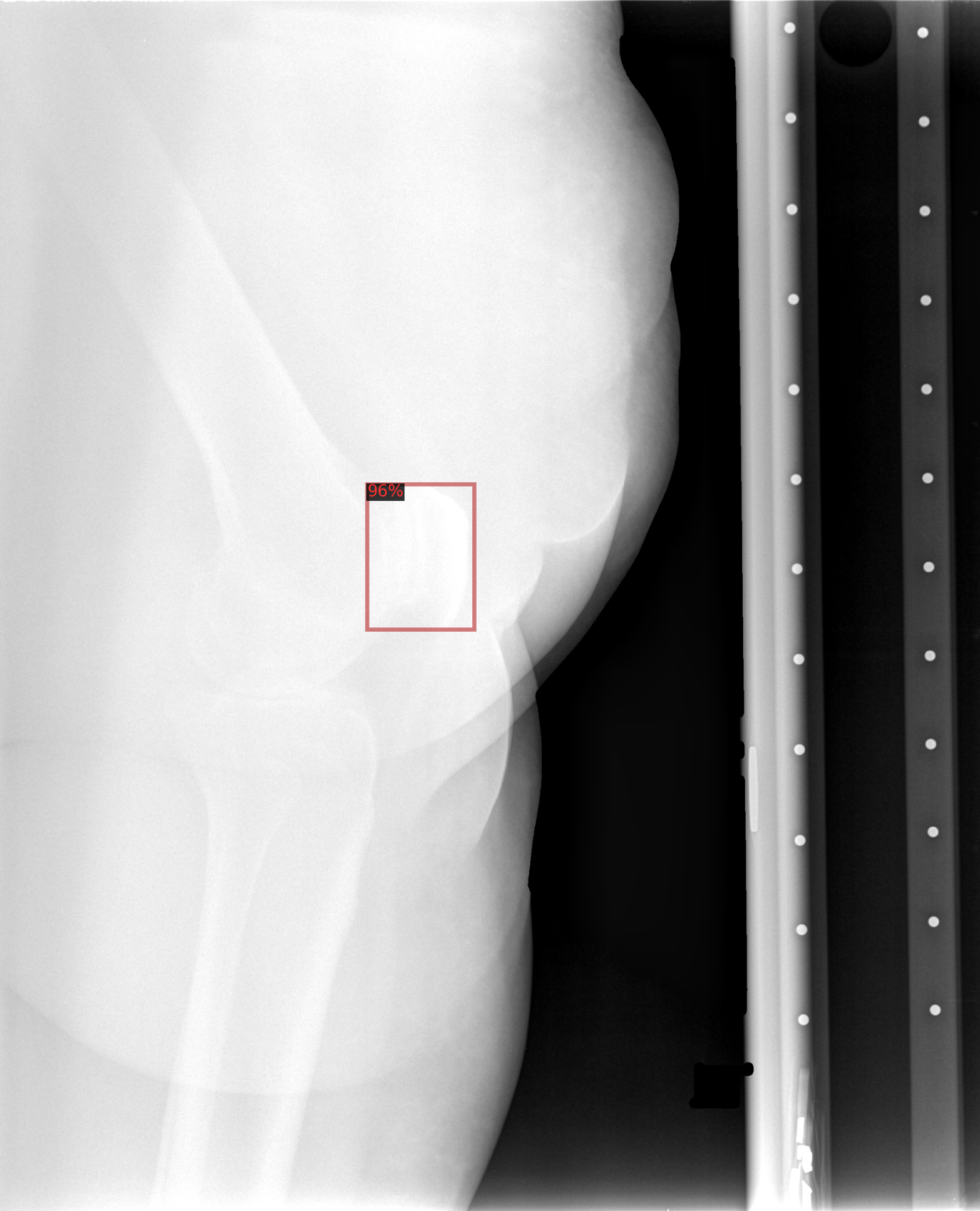}}\hfill
\subfloat[]{\includegraphics[width = 0.7\linewidth, trim={2cm 10cm 8cm 10cm},clip]{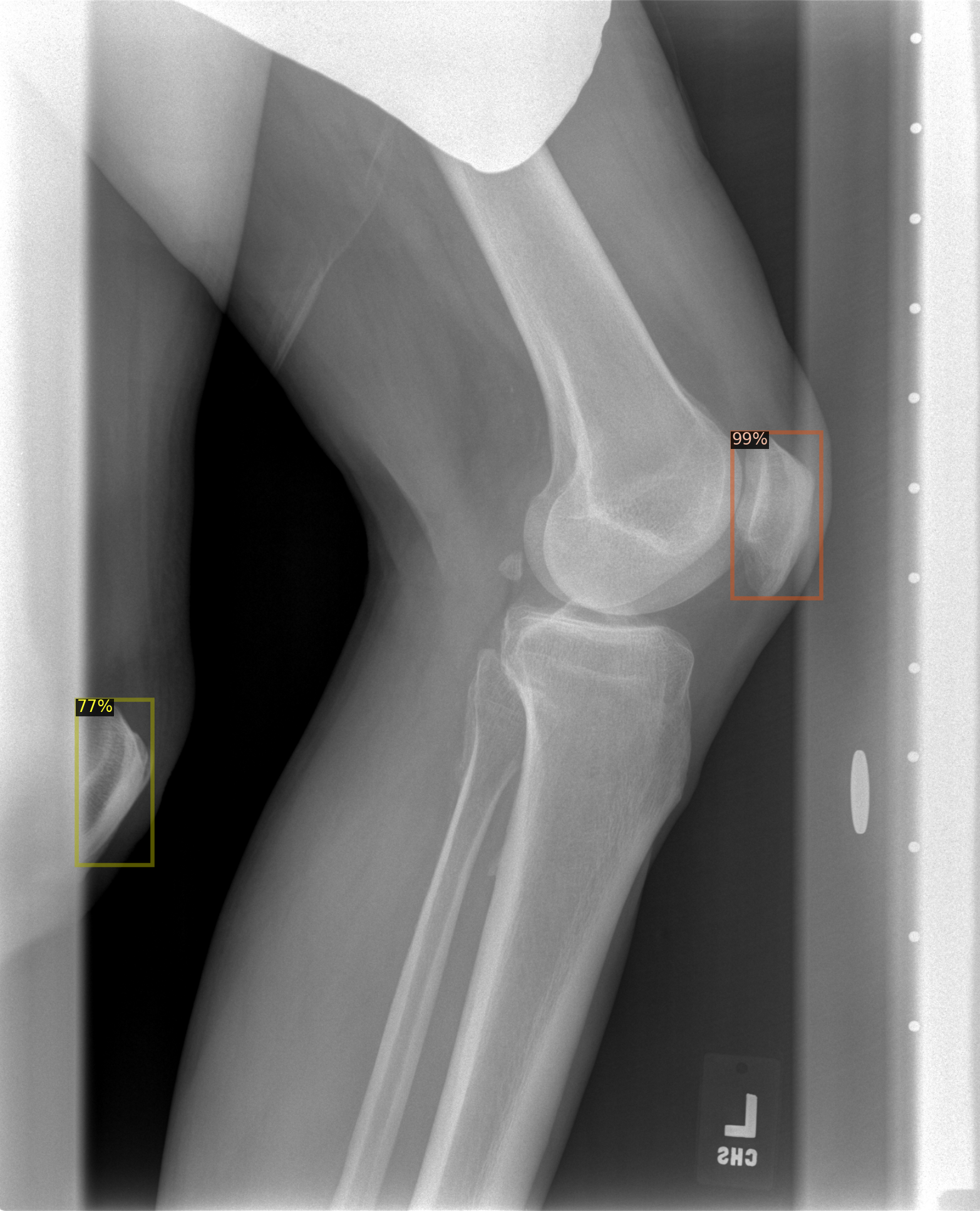}}

\caption{ This figure demonstrates regions found by our patella ROI detector. In (a) patella ROI detected with a high confidence level (96\%) in a challenging sample,  (b) both left and right patella were visible where the detector found both ROIs but since we applied a minimum threshold of 90\% certainty the correct ROI was selected. Images were trimmed from the sides for visibility reasons. Best viewed on screen.}
\label{fig: supp_regions}
\end{figure}

\begin{figure}[!ht]
\subfloat[]{\includegraphics[width = 0.7\linewidth, trim={8cm 12cm 4cm 10cm},clip]{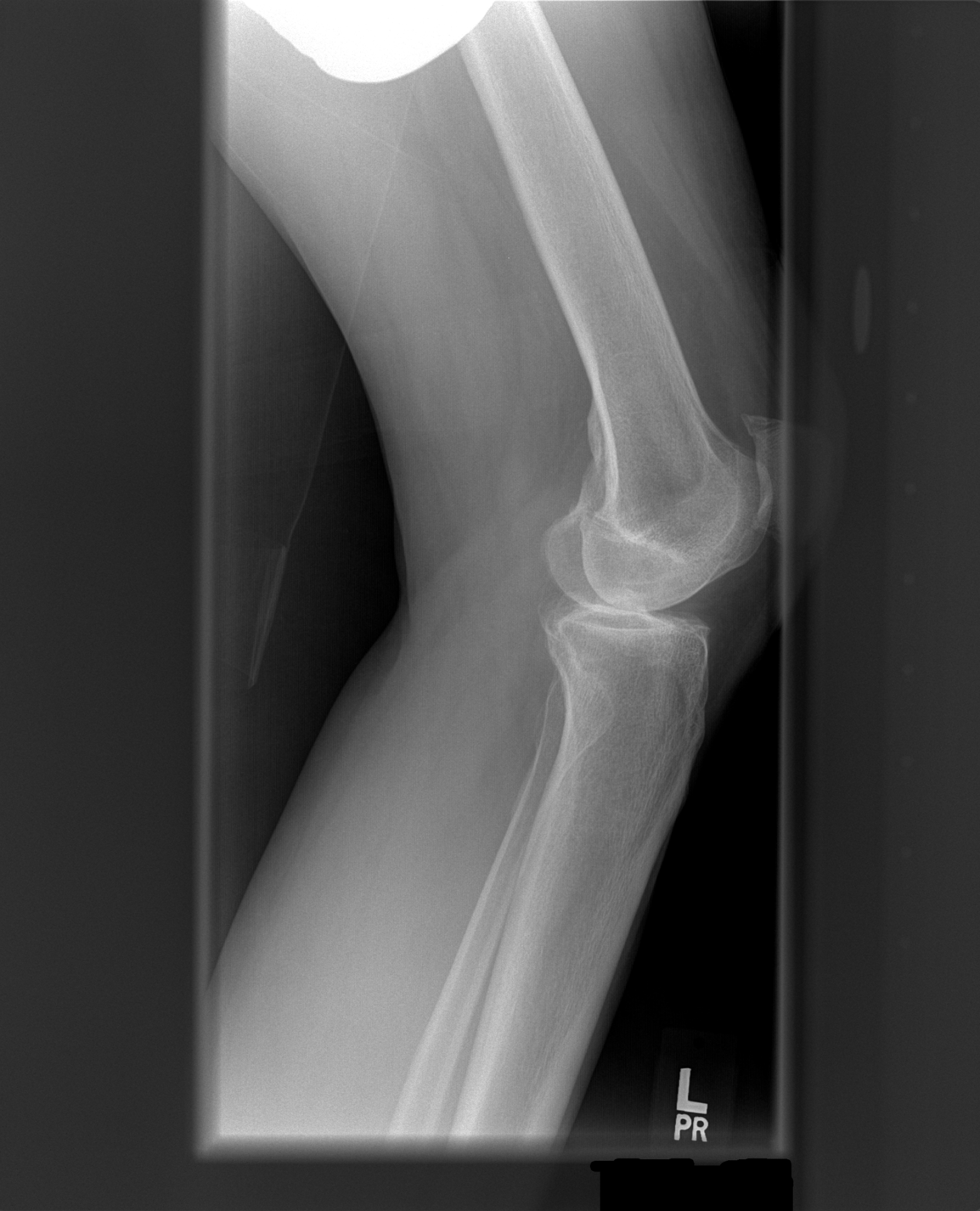}}\hfill
\subfloat[]{\includegraphics[width = 0.7\linewidth, trim={5cm 12cm 8cm 12cm},clip ]{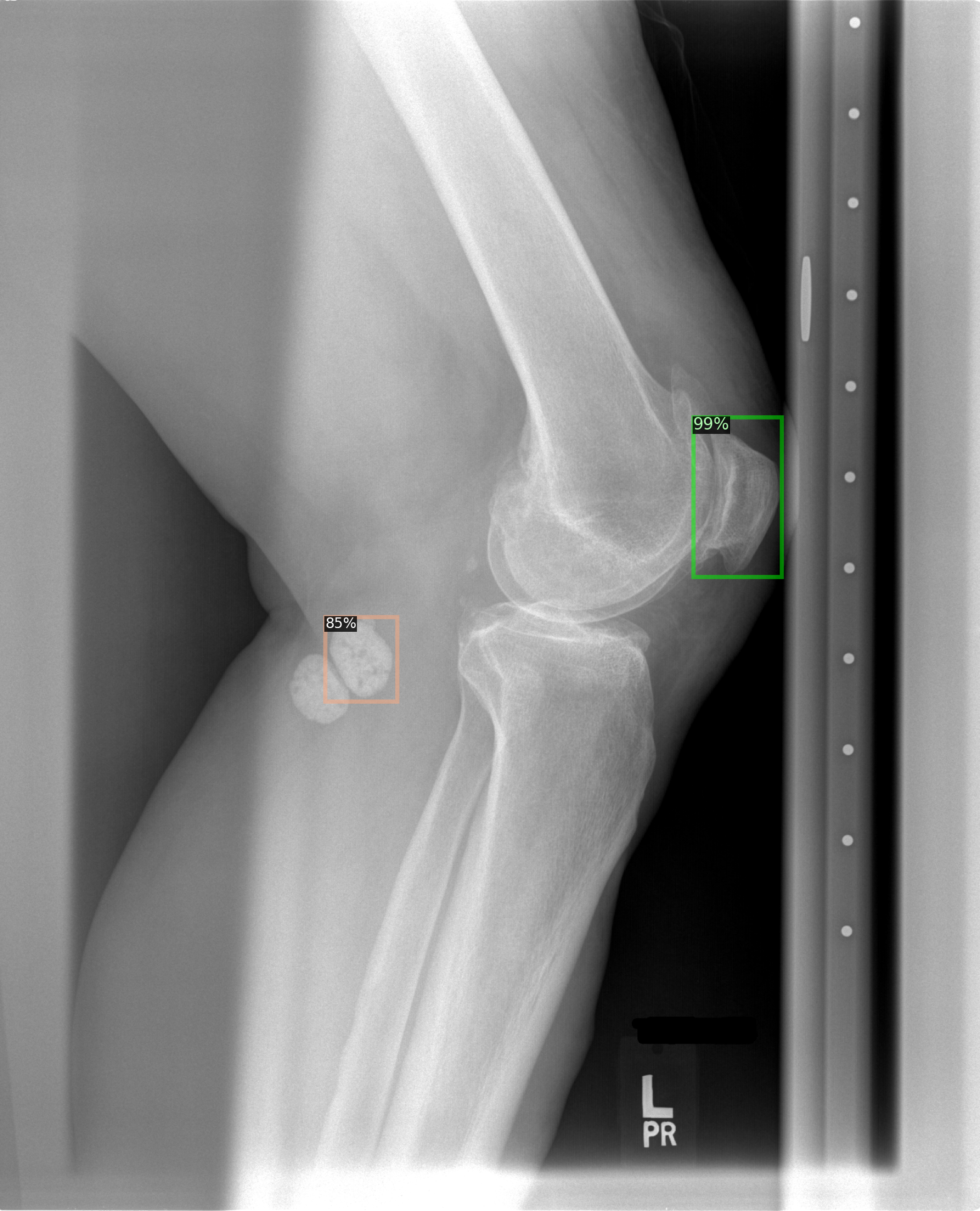}}\hfill

\caption{ In (a) patella ROI was not detected from this radiograph (only 31 patella ROI out of 18467 knees were missed),  (b) non-patella region was detected with a relatively low confidence, again certainty threshold provided the desired result. Images were trimmed from the sides for visibility reasons. Best viewed on screen.}
\label{fig: supp_regions2}
\end{figure}

\begin{figure}[!ht]
\begin{center}
\subfloat[]{\includegraphics[width = 0.7\linewidth]{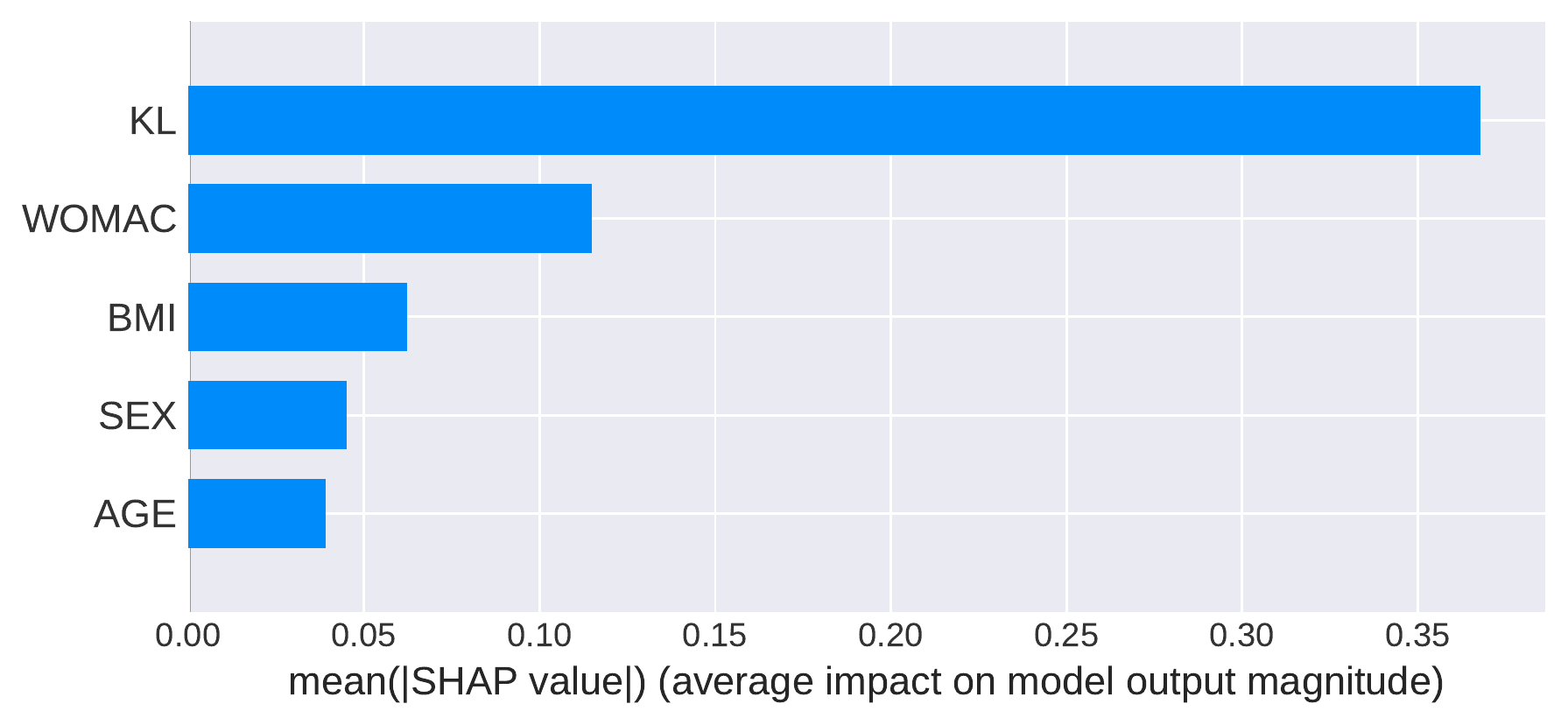}} \\
\subfloat[]{\includegraphics[width = 0.7\linewidth]{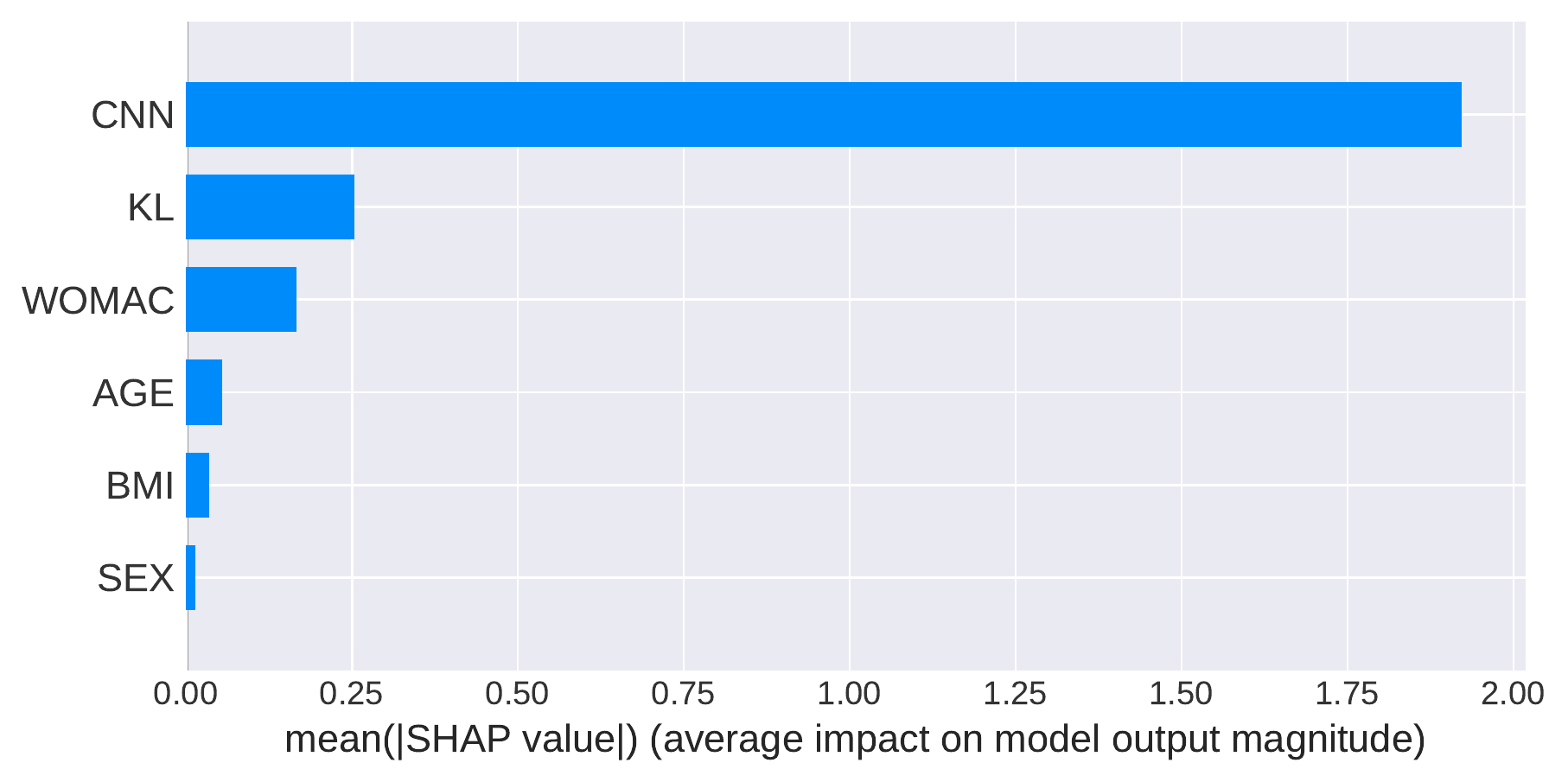}}    
\end{center}

\caption{ SHAP feature importance of (a) model 3  and (b) multi-modal model.}
\label{fig: supp_shap}
\end{figure}

\begin{figure}[!ht]
\subfloat[]{\includegraphics[width = 0.48\linewidth]{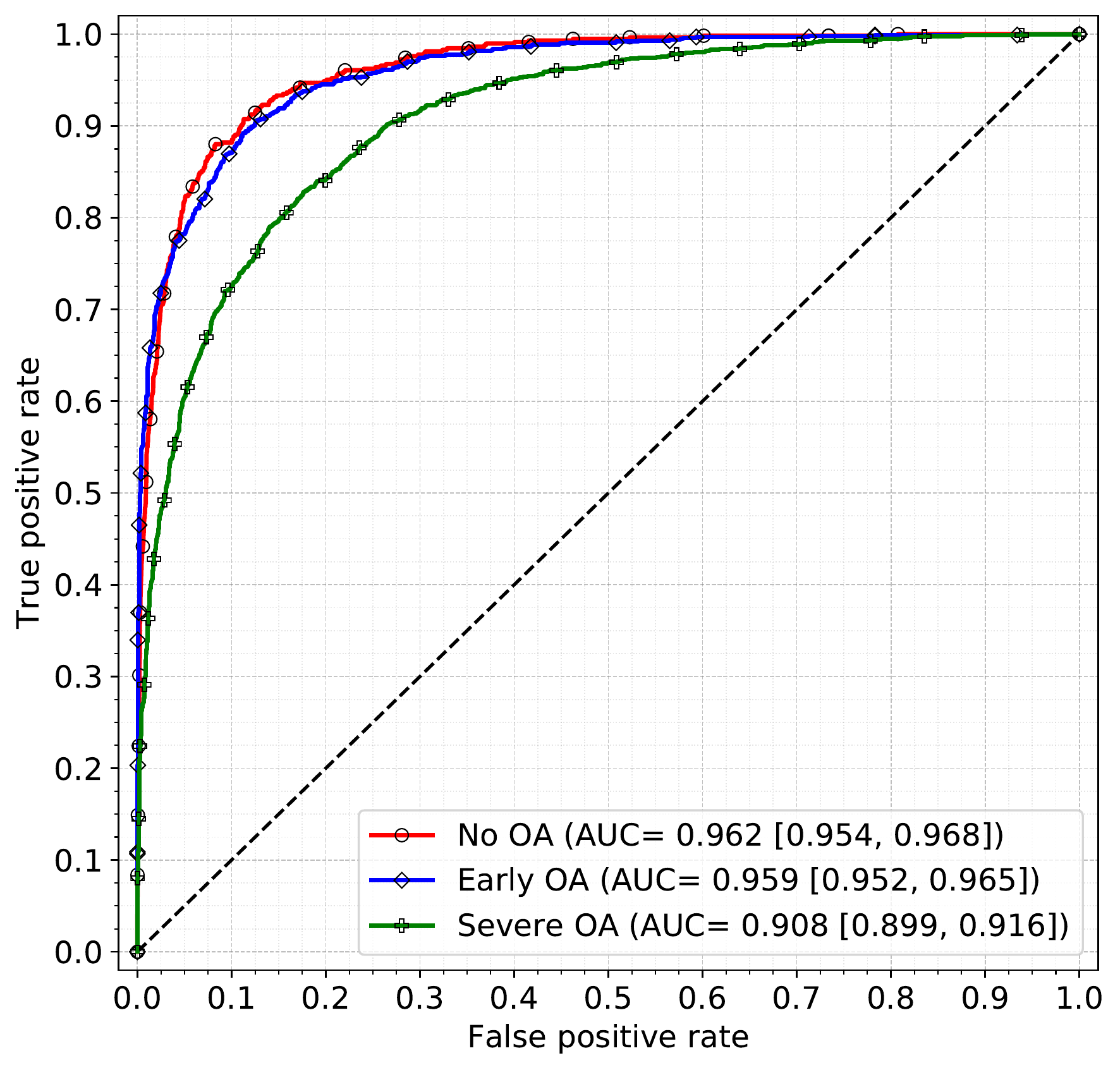}}
\subfloat[]{\includegraphics[width = 0.48\linewidth ]{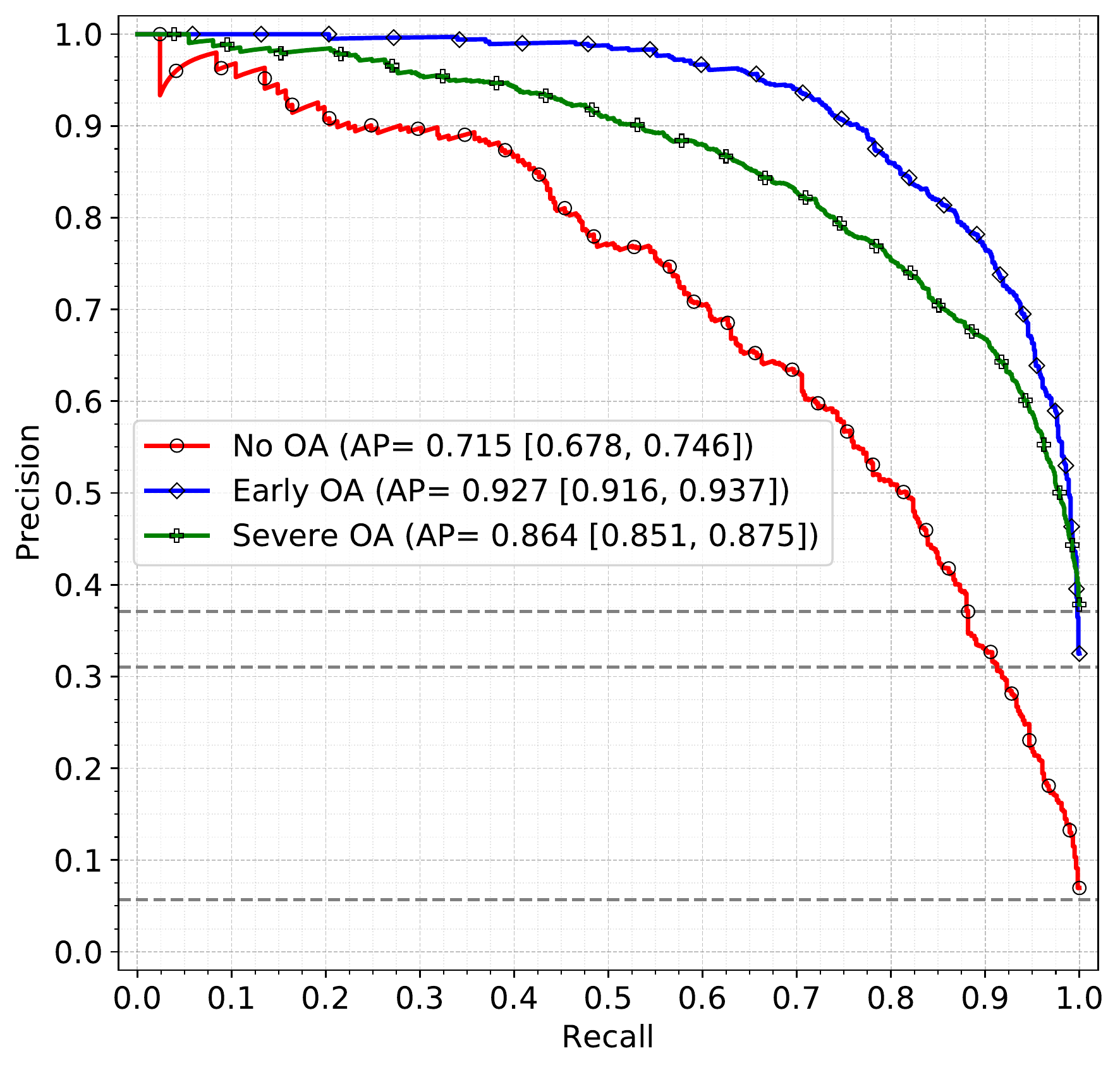}}\\
\subfloat[]{\includegraphics[width = 0.48\linewidth]{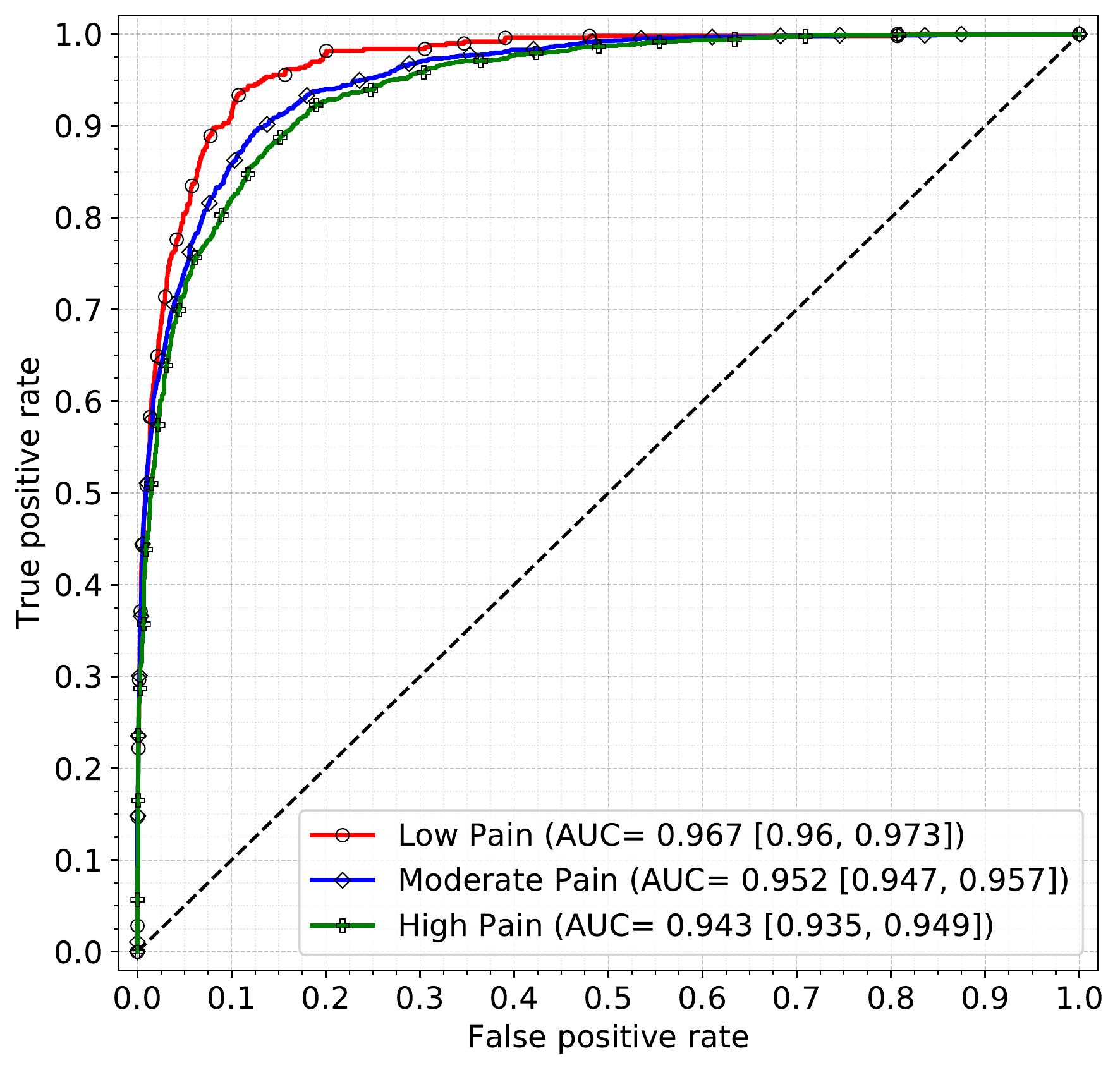}}
\subfloat[]{\includegraphics[width = 0.48\linewidth ]{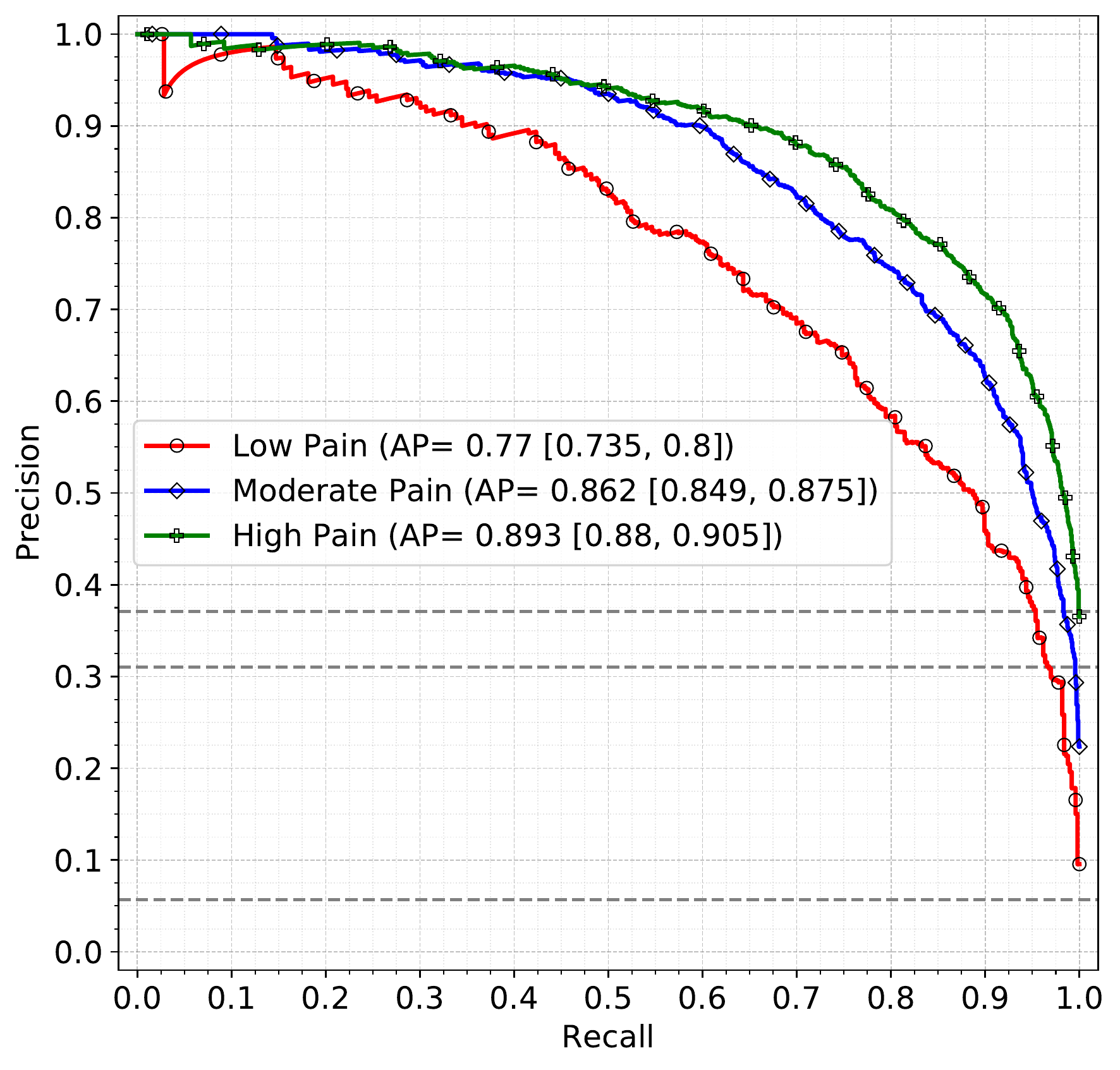}}

\caption{ (a-c) ROC and (b-d) precision-recall curves demonstrating the performance of the CNN model with respect to the stages of tibiofemoral osteoarthritis based on (a-b) Kellgren-Lawrence (KL) grading (No OA : KL0 and KL1, Early OA: KL2, Severe OA : KL3 and KL4 ) and (c-d) the total Western Ontario and McMaster Universities Arthritis Index (WOMAC) Pain score (Low pain: WOMAC pain score $\leq$ $25^{th}$ percentile, Moderate Pain:  WOMAC pain score between $25^{th} - 75^th$ percentile, High pain: $>$ $75^{th}$ percentile).}
\label{fig: supp_groups}
\end{figure}

\begin{figure}[!ht]
\subfloat[]{\includegraphics[width = 0.48\linewidth]{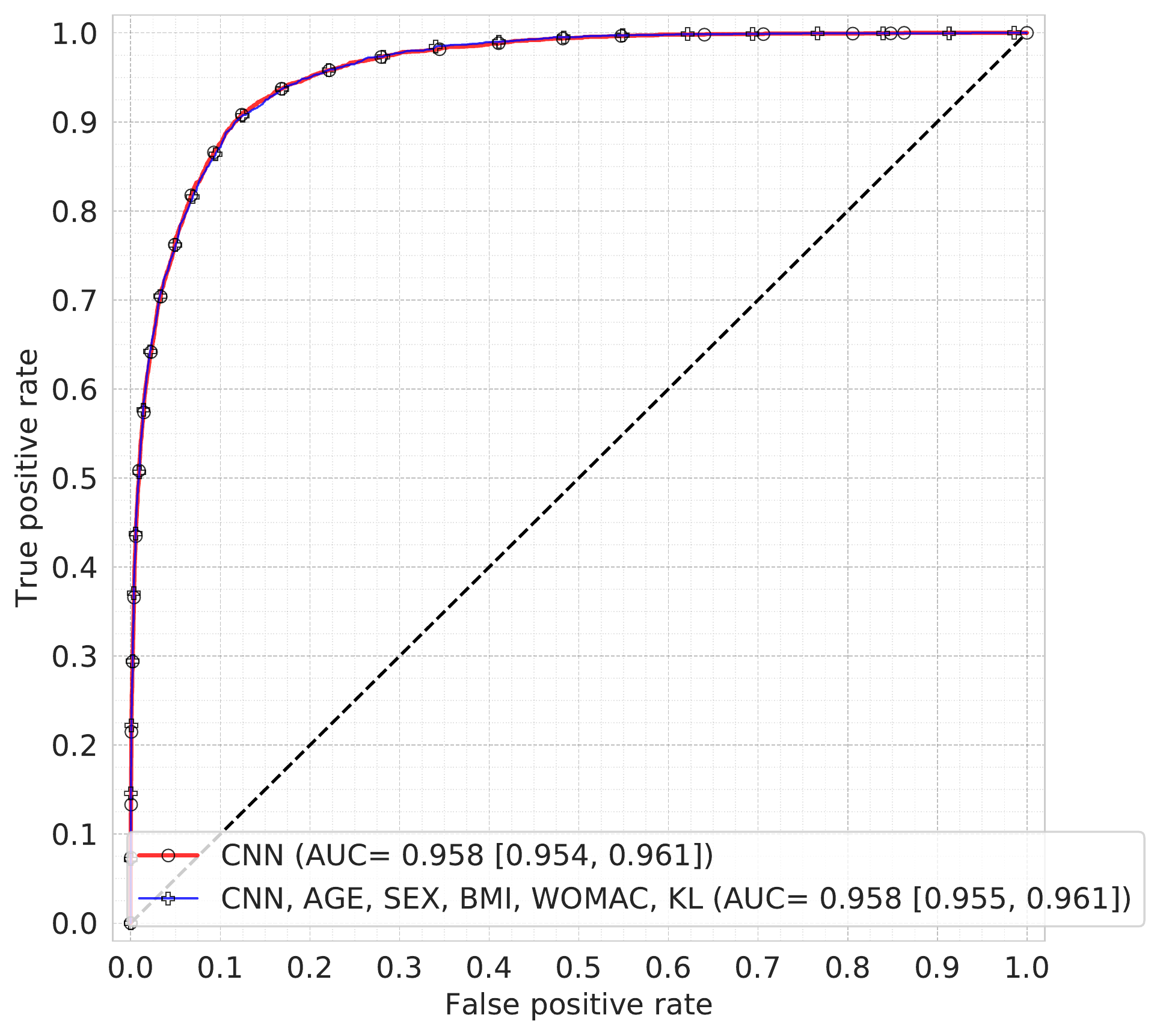}}
\subfloat[]{\includegraphics[width = 0.48\linewidth]{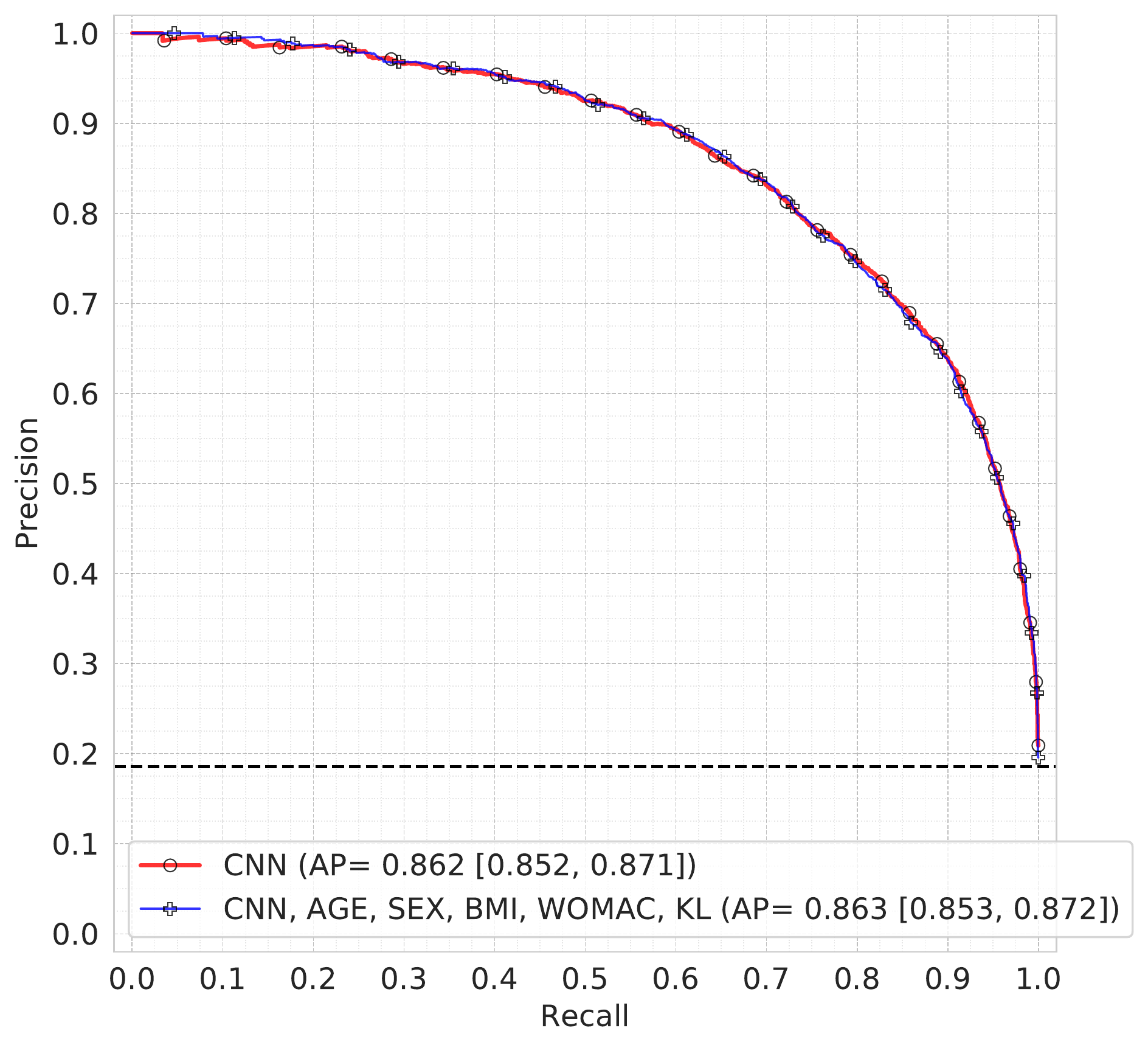}}

\caption{ (a) ROC and (b) precision-recall curves demonstrating the comparison of the deep convolutional neural network (CNN) model with multi-modal method, based on deep CNN, patient characteristics and clinical assessment and Gradient Boosting Machine (GBM) classifier.
The performance of the CNN model and the multi-modal model (CNN+ age, sex, BMI, the total Western Ontario and McMaster Universities Arthritis Index (WOMAC), Kellgren-Lawrence (KL) score was same.}
\label{fig: supp_multimodal}
\end{figure}


\end{document}